\documentclass[journal, 12pt]{IEEEtran}
\usepackage{color,array}
\usepackage[english]{babel}
\usepackage{inputenc}
\usepackage{algorithm}
\usepackage{algorithmicx}
\usepackage{algcompatible}
\usepackage{algpseudocode}
\usepackage{makecell}
\usepackage{cite}
\usepackage{pbox}
\usepackage{amsmath, amssymb}

\usepackage[pdftex]{graphicx}
\usepackage{svg}
\usepackage[figurename=Fig.,font=small,skip=1 pt]{caption}
\graphicspath{{../pdf/}{../jpeg/}}
\graphicspath{ {./img/}}
\graphicspath{ {./img1/}}
\DeclareGraphicsExtensions{.pdf,.jpeg,.png}
\usepackage{epstopdf}
\usepackage{multirow}
\usepackage{accents}
\newcommand{\justified}{%
  \rightskip\z@skip%
  \leftskip\z@skip}
\usepackage{multicol}
\usepackage{enumerate}
\usepackage{array}
\usepackage{color,soul}
\usepackage{pbox}

\usepackage{comment}
\usepackage{url}
\usepackage{stfloats}
\usepackage{amssymb}

\usepackage{setspace}
\usepackage{amsmath}
\usepackage{float}
\usepackage{verbatim} 
\usepackage{color, colortbl}
\definecolor{Gray}{gray}{0.92}
\usepackage{arydshln}
\usepackage{mathrsfs}

\begin{document}



\title{A Lightweight Transformer with Phase-Only Cross-Attention  for Illumination-Invariant Biometric Authentication}



\author{Arun K. Sharma, Shubhobrata Bhattacharya, Motahar Reza, and Bishakh Bhattacharya
      
\thanks{Arun K. Sharma and Bishakh Bhattacharya are with Department of Mechanical Engineering, Indian Institute of Technology, Kanpur, India (e-mail: arnksh@iitk.ac.in and bishakh@iitk.ac.in). \\
Shubhobrata Bhattacharya is with Advanced Technology Development Centre, Indian Institute of Technology, Kharagpur, India (e-mail: emailshubho@gmail.com).\\
Motahar Reza is with Department of Mathematics, School of Science, GITAM Deemed to be University, Hyderabad, India.
(e-mail: mreza@gitam.edu).}
}

\maketitle

\begin{abstract}
Traditional biometric systems have encountered significant setbacks due to various unavoidable factors, for example, wearing of face masks in face recognition-based biometrics and hygiene concerns in fingerprint-based biometrics. This paper proposes a novel lightweight vision transformer with phase-only cross-attention (POC-ViT) using dual biometric traits of forehead and periocular portions of the face, capable of performing well even with face masks and without any physical touch, offering a promising alternative to traditional methods. The POC-ViT framework is designed to handle two biometric traits and to capture inter-dependencies in terms of relative structural patterns. Each channel consists of a Cross-Attention using phase-only correlation (POC) that captures both their individual and correlated structural patterns. The computation of cross-attention using POC extracts the phase correlation in the spatial features. Therefore, it is robust against variations in resolution and intensity, as well as illumination changes in the input images. The lightweight model is suitable for edge device deployment. The performance of the proposed framework was successfully demonstrated using the Forehead Subcutaneous Vein Pattern and Periocular Biometric Pattern (FSVP-PBP) database, having 350 subjects. The POC-ViT framework outperformed state-of-the-art methods with an outstanding classification accuracy of $98.8\%$ with the dual biometric traits.
\end{abstract}
\begin{IEEEkeywords}
Cross-spectral Vision Transformer, Biometric Authentication, Forehead Subcutaneous Vein Pattern, Periocular Pattern, Multihead Cross-attention.
\end{IEEEkeywords}
\IEEEpeerreviewmaketitle

\section{Introduction}
\label{intro}
In the evolving landscape of biometric identification, the quest for more secure, reliable, non-contact, and non-invasive technologies has become paramount, especially in the wake of global challenges such as the COVID-19 pandemic \cite{gomez2022biometrics}. Traditional biometrics like face recognition \cite{wang2023masked}, fingerprint recognition \cite{10056846}, iris recognition \cite{tpami2, adamovic2020efficient}, hand palm vein \cite{lee2012novel}, hand dorsal vein \cite{joardar2014real}, wrist vein \cite{pascual2010capturing}, finger vein \cite{das2018convolutional}, and multimodal biometrics \cite{tpami1} etc has inherent limitations and vulnerabilities that undermine their efficacy and user acceptance. Among these, facial and fingerprint recognition systems stand out as the most prevalent in commercial and semi-commercial applications due to their non-intrusiveness and ease of use. However, the mask-wearing norms and hygiene concerns have significantly impacted their reliability and acceptance \cite{gomez2022biometrics}, propelling the need for alternative biometric solutions that can offer comparable or superior levels of identification accuracy with no drawbacks associated with full-face or touch-based systems.

In recent decades, the forehead vein pattern recognition \cite{bhattacharya2022portable} has emerged as a promising candidate in this context, offering a novel approach to biometric identification that leverages the unique physiological features of the human forehead. This method not only aligns with the requirements for non-intrusive and touchless hygienic biometric systems but also addresses the challenges posed by facial occlusions. The rationale behind focusing on the forehead as a biometric trait lies in its inherent advantages; for example, the forehead remains largely unaffected by facial expressions and is less likely to be obscured by accessories or hair, providing a stable region for identification. Secondly, the area is easily accessible for imaging, even in the presence of face masks, making it an ideal candidate for recognition in today's real-world scenario.

Bhattacharya \textit{et. al.} \cite{bhattacharya2022portable} has attempted to explore the viability of a contactless and real-time system based on the forehead subcutaneous vein pattern and periocular biometric pattern (FSVP-PBP) as an alternative biometric identification method. It uses a convolutional neural network-based feature extraction for the vein pattern and the periocular pattern independently, followed by feature concatenation for final classification. The convolutional-based feature extraction method relies on kernel-based spatial features in the input images, which are highly dependent on the resolution of the input images. Therefore, it may fail to capture the subject-specific structural patterns in the forehead image. Furthermore, if the depth of CNN-based models is significantly increased for better performance \cite{das2018convolutional}, the computational complexity increases so high that it raises a question of their deployability for real-time applications.  Therefore, to improve the robustness against the illumination and the image resolution, this paper presents an innovative lightweight vision transformer architecture called POC-ViT with a novel phase-only correlation-based cross-attention (POC-CA) to capture the subject-specific structural pattern from the vein pattern images.
The highlights of the contributions of this work are listed below:
\begin{enumerate}[i)]
    \item An adaptive Tann-Triggs algorithm has been formulated for the preprocessing of the NIR camera-based images to make the vein patterns prominently visible.
    \item An innovative dual-architectural framework of POC-ViT is formulated to capture the subject-specific structural information from the raw images of the forehead subcutaneous portion and the periocular portion of the subject. The proposed framework of POC-ViT consists of multihead cross-spectral attention that leverages the concept of Phase-only correlation (POC) to capture the relative orientation of vein patterns for subject-specific feature extraction.
    \item The performance of the proposed framework was successfully tested on an extended FSVP-PBP database of 350 subjects under two conditions: (i) with the raw FSVP-PBP dataset (no preprocessing) and (ii) with the FSVP-PBP dataset preprocessed using the proposed adaptive Tann-Triggs algorithm. The evaluation results prove the superiority of the proposed framework over state-of-the-art algorithms and make it a new benchmark in the domain of non-touch and secure biometric authentication.
\end{enumerate}

The remaining parts of the paper are organized as follows. Section \ref{sec:proposed} describes the proposed POC-ViT framework. Section \ref{sec:experi} presents the effectiveness of the proposed framework and its comparison with state-of-the-art methods on the FSVP-PBP database of 350 subjects. Finally, Section \ref{sec:conclusion} concludes the work.

\section{Literature Review}
\label{LiteratureReview}
\subsection{Review of Vein Pattern Recognition (VPR) based on Approaches}
\label{LiteratureReviewI}
\subsubsection{Graph-based VPR}
\label{GraphBasedVeinPatternRecognition}
The graph-based VPR relies on feature extraction based on geometrical properties and spatial structures for pattern recognition. The line-like and/or curve-like models: the repeated line tracking method \cite{miura2004feature}, the maximum curvature point \cite{miura2007extraction}, mean curvature \cite{song2011finger}, and Gabor filter \cite{zhang2013finger} have been successfully adopted for robust and discriminative feature extraction. Additionally, statistical analysis tools like principal curvature \cite{choi2009finger} and vector grams of maximal intra-neighbor difference \cite{kang2012vein} have also been applied for discriminative feature extraction. The low contrast distribution and uneven illumination cause these methods to be very unreliable and inaccurate.

\subsubsection{Deep Learning-based VPR}
\label{generalliteraturesurvey}
Various researchers have explored the performance of the convolutional neural network (CNN)-based deep-learning models for VPR \cite{radzi2016finger, itqan2016user, wang2017hand, das2018convolutional}. The CNN-based methods have outperformed the manually crafted feature methods in the image classification tasks, however, their performances are very much affected due to the scarcity of sufficient vein pattern datasets. Radzi et al. \cite{radzi2016finger} introduced a simplified four-layer CNN with an integrated convolutional-sampling structure, trained on a limited dataset. This network, also used in finger vein recognition \cite{itqan2016user}, yielded significantly lower performance than traditional models, highlighting its unsuitability for small datasets. To enhance CNN-based vein pattern recognition, modifying the network architecture for specific tasks \cite{wang2017hand, vgg16, das2018convolutional} or developing quality-aware models \cite{qin2017deep} have been explored, showing significant improvements. Some studies propose using a pre-trained CNN \cite{zhong2018palm} and \cite{wang2018minutiae} as a feature extractor, drawing features from the fully connected layer or using minutiae-based pooling. However, these methods suffer from biases due to differences between training and target datasets. Noh \textit{et. al.} \cite{noh2020finger} presented a DenseNet model with a score level feature fusion of the model output the texture and the shape images of multimodal databases taken from Shandong University homologous multi-modal finger-vein (SUHM-FV) and Hong Kong Polytech University finger image (HKPolyU-FI) database. Li \textit{et.al.} \cite{li2021local} proposed a local discriminant coding-based CNN model for feature extraction in multimodal finger images. The authors reported a very high accuracy of 99.9\% on the tri-model (fingerprint (FP), finger-vein (FV), and finger knuckle-print (FKP)) finger image dataset; however, the approach relies on the contact-based multiple views of fingerprint images. Also, Daas \textit{et. al.} \cite{daas2020multimodal} presented a fusion-based deep learning model with FV and FKP database for the development of a multimodal biometric system. They created a multimodal dataset using the two databases: Shandong University homologous multi-modal finger-vein and Hong Kong Polytech University FKP database (SUHM-FV + PolyU FKP).

\subsubsection{Attention-based VPR} 
\label{generalliteraturesurvey2}
In recent advances in vein-based biometric recognition, various authors have investigated the use of attention-based modern computer vision models \cite{FV-ViT, FV-LT, VPCFormer, blstm, albano2024explainable}. Li and Zhang \cite{FV-ViT} introduced FV-ViT, a variant of vision transformer with regularization added in the MLP head, and reported outstanding performance with EER of 0.042\% on FV-USM and 1.033\% on SDUMLA-HMT datasets. Qin \textit{et. al.} \cite{FV-LT} introduces a local attention transformer for a full-view finger-vein pattern recognition system that relies on $360^{\circ}$ views captured by a rotating LED-based image acquisition system. Further, Qin \textit{et. al.} \cite{blstm} introduced an attention bidirectional LSTM-based temporal-spatial vein transformer that used LSTM with attention and local attention in the ViT model for temporal and spatial feature extraction in multi-view finger vein patterns. In another work on multi-view finger vein recognition, Zhao \textit{et. al.} \cite{VPCFormer} introduced a Transformer-based model with a vein pattern attention module and an integrative feed-forward network that uses a vein mask to extract the features from multi-view efficiently. Further, the study by Albano et. al. \cite{albano2024explainable} leverages explainable AI to enhance transparency in wrist-vein recognition using Vision Transformer (ViT), demonstrating improved verification performance by focusing on key vein patterns. 

\subsection{Review of VPR based on Modality}
\label{LiteratureReviewII}
\subsubsection{Finger vein Pattern}
\label{FinureVeinPattern}
Over the past ten years, there have been remarkable improvements in the field of single-view finger vein recognition technology. These advancements have been fueled by the noticeable difference in appearance between the finger veins and their surroundings in images. Techniques such as Local Maximum Curvature (LMC) \cite{miura2007extraction}, Wide Line Detector (WLD) \cite{huang2010finger}, and Enhanced Maximum Curvature (EMC) \cite{syarif2017enhanced} have been reported.  To enhance feature extraction, SSP-DBFL \cite{zhao2023neglected} was introduced, combining two types of input features for more effective learning. With the rise of deep learning in the realm of computer vision, several scholars have tailored these techniques for better finger vein recognition. One notable example is FV-GAN \cite{yang2019fv}, which employs Generative Adversarial Networks (GANs). Similarly, Song et al. \cite{song2022eifnet} introduced EIFNet, a model that excels in merging both implicit and explicit features for improved discrimination. Additionally, inspired by the achievements of Vision Transformer (ViT) \cite{dosovitskiy2020image}, Huang et al. created the Finger Vein Transformer (FVT) \cite{huang2022fvt}, which uses a pyramid structure to extract features at multiple levels. Further, Haung \textit{et. al.} \cite{huang2023multimodal} introduced a finger asymmetric backbone network to obtain discriminating intra-model features for multimodal figure databases, for example, fingerprint and finger vein pattern recognition.

\subsubsection{Hand Dorsal vein Pattern}
\label{HandDorsalVeinPattern}
In 1985, Joseph Rice invented dorsal hand vein (DHV) pattern recognition, securing the first U.S. patent for this technology \cite{rice1987apparatus}. In \cite{cross1995thermographic}, Cross and Smith advanced the technology with thermographic imaging for biometrics. Subsequent advancements included Lin et al.'s thermal image-based recognition in \cite{lin2004biometric} and Huang et al.'s high-accuracy local feature matching in \cite{huang2014hand}. In \cite{li2016comparative}, Li et al. applied deep learning to DHV recognition, demonstrating the effectiveness of CNNs. Wang et al. in \cite{wang2017quality} proposed an imaging device to create a dorsal hand vein pattern (DHVP) database, employing DNN models for identity and gender recognition. Advancements continued with Gupta et al.'s multimodal biometric system in \cite{gupta2018multibiometric}, integrating DHV with other biometrics. Further, Egorov et al. \cite{10510055} proposed the use of scale-invariant feature transform (SIFT) for the feature extraction for DHV pattern recognition.

\subsubsection{Hand Palm vein Pattern}
\label{HandPalm}
Numerous studies have been reported for the palm vein recognition \cite{kang2014contactless, lee2015palm, ma2017palm, bharathi2019biometric, horng2021recognizing}. Kang et al. \cite{kang2014contactless} developed a contactless identification method using Mutual foreground LBP (MF-LBP) for texture matching and validated it on the CASIA multispectral palmprint image dataset (CASIA-MPID). Lee et al. \cite{lee2015palm} introduced the modified 2D$^{2}$ LDA method, enhancing classical LDA with matrix representations to address singularity issues in high-dimensional data, and reducing computational demands. Xin et al. \cite{ma2017palm} merged local Gabor histograms with LGPDP and LGXP to improve data discrimination and optimized Gabor filter parameters through image subdivision. Bharathi et al. \cite{bharathi2019biometric} focuses on a multimodal biometric system using vascular patterns of the hand, such as finger vein and palm vein images. Recently, Horng et al. \cite{horng2021recognizing} explored palm vein recognition in smartphones using RGB images, marking a significant step toward mobile applications.

\subsubsection{Forehead vein Pattern}
\label{ForeHeadVein}
Forehead biometrics emerges as a promising, non-intrusive identification method, offering advantages in hygiene and usability, especially under mask-wearing conditions. Recent studies highlight its potential for high accuracy and reliability, leveraging unique forehead features, despite challenges in environmental variability and algorithm development \cite{bhattacharya2022portable}. This method represents a viable alternative to traditional biometrics, adapting to contemporary needs.

Apart from these modalities, Santos \textit{et. al.} \cite{santos2015fusing} presented a cross-sensor based biometric system that provided a cross-sensor iris and periocular (CSIP) database having iris and periocular patterns. The results demonstrated the biometric authentication with an equal error rate of 0.145.
\section{Proposed Methodology} 
\begin{figure*}[!ht]
\centering
\includegraphics[width=0.75\textwidth]{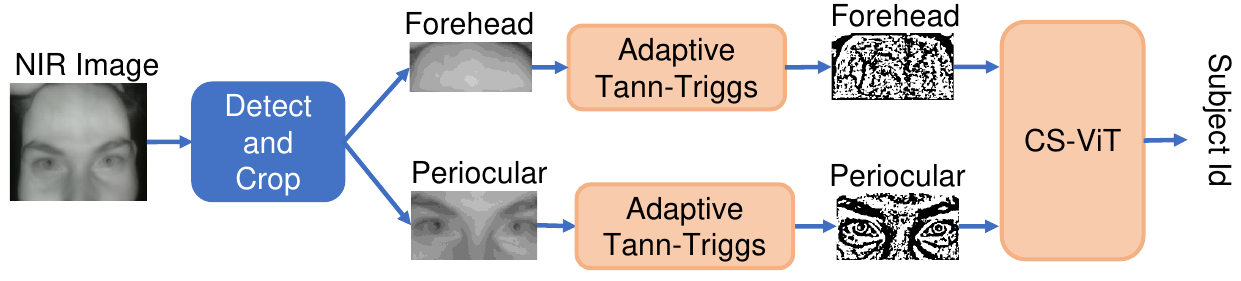}
\caption{Complete block diagram of the proposed framework. ATT: Adaptive Tann-Triggs algorithm.}
\label{fig:overAll}
\end{figure*}

The overall block diagram for the proposed methodology is illustrated in Fig. \ref{fig:overAll}. The NIR camera images are cropped and separated into the forehead and the periocular portions using the \textit{Haar Cascade Algorithm} provided by \textit{OpenCV}. The cropped images are passed through an adaptive Tann-Triggs (ATT) algorithm to transform them to have prominently visible patterns. Now, the processed images of the forehead and the periocular portions are fed to the proposed POC-ViT framework for subject identification.

\subsection{Adaptive Tann-Triggs Algorithm}
\label{preprocessing}
Tan and Triggs \cite{tan_trig2010} introduced a chain of preprocessing techniques to eliminate the effect of illumination variations and transform the near-infrared (NIR) camera-based images into prominently visible patterns. The original Tann-Triggs algorithm requires suitable tuning of user-defined parameters to get the output image with prominently visible veins. The tuning of these hyperparameters is a time-consuming process, or sometimes the manually selected parameters may not produce the desired output.  Therefore, this section presents an adaptive Tann-Triggs algorithm that adaptively chooses the hyperparameters based on the image statistics and transforms the NIR image to have prominently visible vein patterns. The chain of the adaptive preprocessing steps is discussed below:

\subsubsection{Adaptive Gamma Correction (AGC)}
\label{gammacorrection}
Adaptive Gamma Correction (AGC) is a gray-level transformation to eliminate the effect of the luminance on the NIR image using the value of $\gamma_a$, calculated adaptively based on the standard deviation $\sigma$ of the normalized image. The gamma-corrected image with adaptive value of $\gamma$ ($I_{agc})$ is given by equation (\ref{eq:gmm}).
\begin{equation}\label{eq:gmm}
    I_{agc}(x,y) = I_{n}^{\gamma}(x,y)
\end{equation}
where, $I_{n}$ is the normalized image and $\gamma = \text{max}\left(0.2, \min(1, 1-\sigma)\right)$ with $\sigma$ be the standard deviation of the normalized image, calculated as $\sigma = \sqrt{\frac{1}{N}\sum_{i=1}^N{(I_n(i)-\mu)^2}}$; $\mu$ = mean pixel value, and $N$ = number of pixels.

\subsubsection{Adaptive Difference of Gaussian (DoG) filter}
\label{dog}
The high-frequency component is further enhanced by applying the difference of Gaussian filters with the adaptively selected kernels $\sigma_0$ and $\sigma_1$, where $\sigma_1$ = 2$\sigma_0$ and $\sigma_0$ is adaptively selected based on the standard deviations of pixel intensities. If $G_0(x,y)$ and $G_1(x,y)$ are the Gaussian blurs for $\alpha_0$ and $\alpha_1$, respectively, the DoG filtered image $I_{DoG}(x,y)$ is obtained as 
$$I_{DoG}(x,y) = G_0(x,y)-G_1(x,y)$$
where, $\alpha_0 = \sqrt{\frac{1}{N}\sum_{i=1}^N{(I_i-\mu)^2}}$; adaptive to the intensity ($I_i$) of the $i^{th}$ pixel and the mean pixel ($\mu$). $N$ be the total number of pixels in the gamma-corrected image. Further details on the computation of Gaussian blur for applying the DoG filter can be found in \cite{tan_trig2010}.

\subsubsection{Adaptive Contrast Equalization (ACE)}
\label{ce}
Now, the uneven/extreme pixel intensity at various portions of the image is suppressed by zero-mean unit variance normalization on the DoG filtered image $I_{DoG}(x,y)$, followed by applying a non-linear dynamic truncation as shown in equations (\ref{eq:zmuv}) and (\ref{eq:nlt}). 
\begin{equation}\label{eq:zmuv}
    I_z(x,y) = \frac{I_{DoG}(x,y)-\mu_{DoG}}{\sigma_{DoG}+\epsilon}
\end{equation}
\begin{equation}\label{eq:nlt}
    I_{ACE}(x,y) = \begin{cases} 
-\tau, & \text{for } I_z(x, y) < -\tau, \\
I_z(x, y), & \text{for } -\tau \leq I_z(x, y) \leq \tau, \\
\tau, & \text{for } I_z(x, y) > \tau,
\end{cases}
\end{equation}
where, $\tau$ is adaptively selected based on the input image distribution as $\tau = 1.5\sigma_{DoG}$; $\sigma_{DoG}$ and $\mu_{DoG}$ are the standard deviation and mean of the $I_{DoG}(x,y)$.

\begin{figure*}[!ht]
\centering
\includegraphics[width=0.8\textwidth]{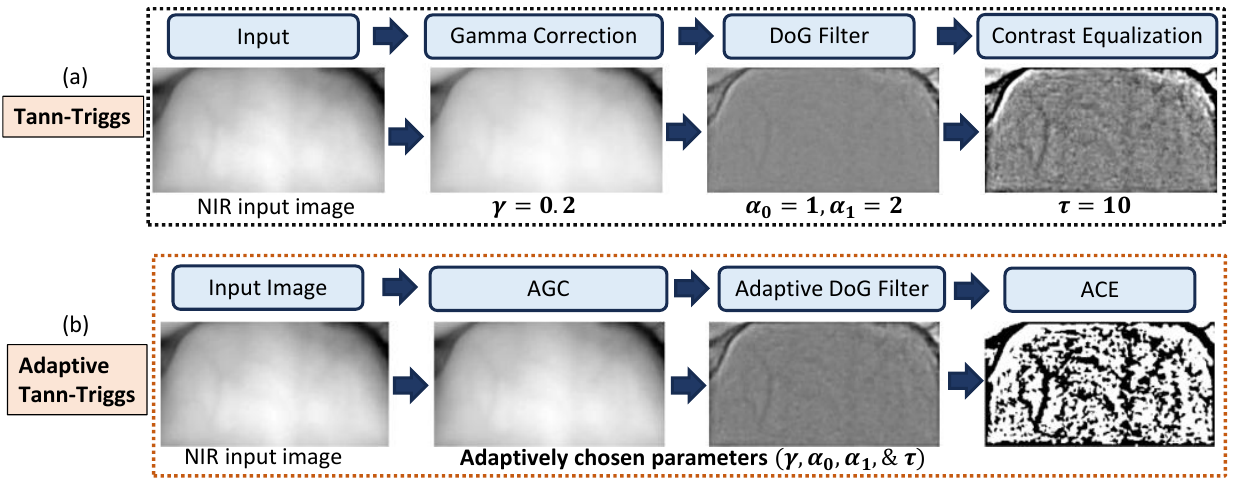}
\caption{Comparing the effect of image statistics-based adaptive parameter selection image preprocessing by Tann-Triggs algorithm: (a) Standard Tann-Triggs with user-defined parameters \cite{tan_trig2010} and (b) proposed adaptive Tann-Triggs algorithm with adaptively chosen parameters based on image statistics.}
\label{fig:demo_att}
\end{figure*}

The $I_{CE}(x,y)$ is further normalized to obtain the final enhanced image. The benefit of adaptively selected parameters like $\gamma$ exponent for gamma correction, $\alpha_0$ and $\alpha_1$ for DoG filtering,  and $\tau$ for CE is illustrated in Fig. \ref{fig:demo_att} by comparing the outcomes with the default suggested values in the original paper \cite{tan_trig2010}.

\subsection{Proposed POC-ViT }\label{sec:proposed} 
\begin{figure*}[!ht]
\centering
\includegraphics[width=0.8\textwidth]{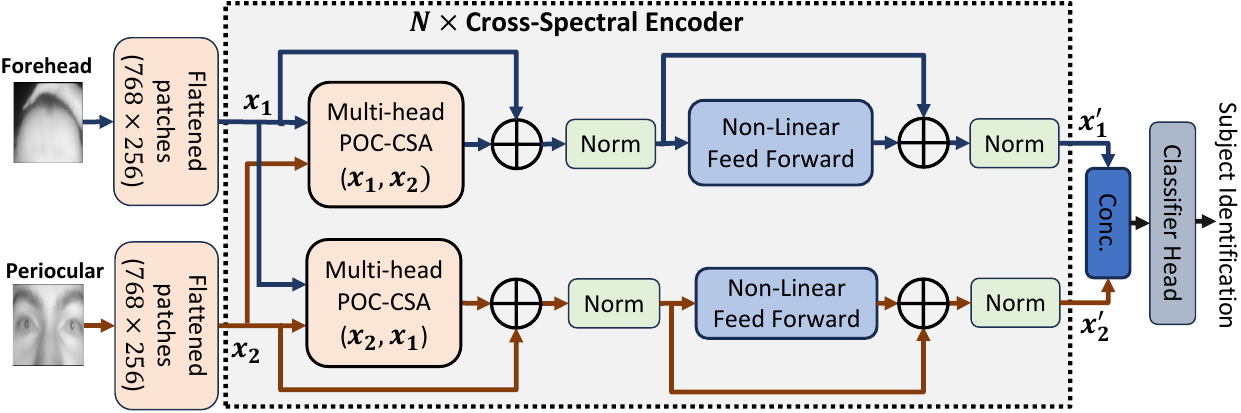}
\caption{Proposed framework of cross-spectral vision transformer (POC-ViT) architecture.}
\label{fig:proposed}
\end{figure*}

Vision Transformer (ViT) introduced by Dosovitskiy et al. \cite{vit} has shown outstanding performance for computer vision applications. The attention mechanism in the ViT provides the capability of capturing the long dependencies in the flattened sequence of the input image in the form of patches. The proposed framework of \textbf{P}hase-only Correlation-based \textbf{C}ross-Attention for \textbf{Vi}sion \textbf{T}ransformer (\textbf{POC-ViT}) comprises N numbers of cross-attention encoder blocks similar to ViT, but each encoder consists of a dual-channel multi-head POC-based cross-attention (POC-CA). The dual-channel of multi-head POC-CA has been specifically designed for learning the person-specific signature using dual biometric traits. The Phase-only correlation (POC) \cite{kuglin1975phase} serves as a powerful tool for capturing the alignment or structural similarity between two biometric traits in the frequency domain. The schematic diagram of the proposed framework has been shown in Fig. \ref{fig:proposed}. The two input channels receive flattened patches of NIR camera-based images of the forehead and the periocular portion of faces. Further detailed descriptions of each step are provided in the following sub-sections.

\begin{figure}[!ht]
\centering
\includegraphics[width=0.8\columnwidth]{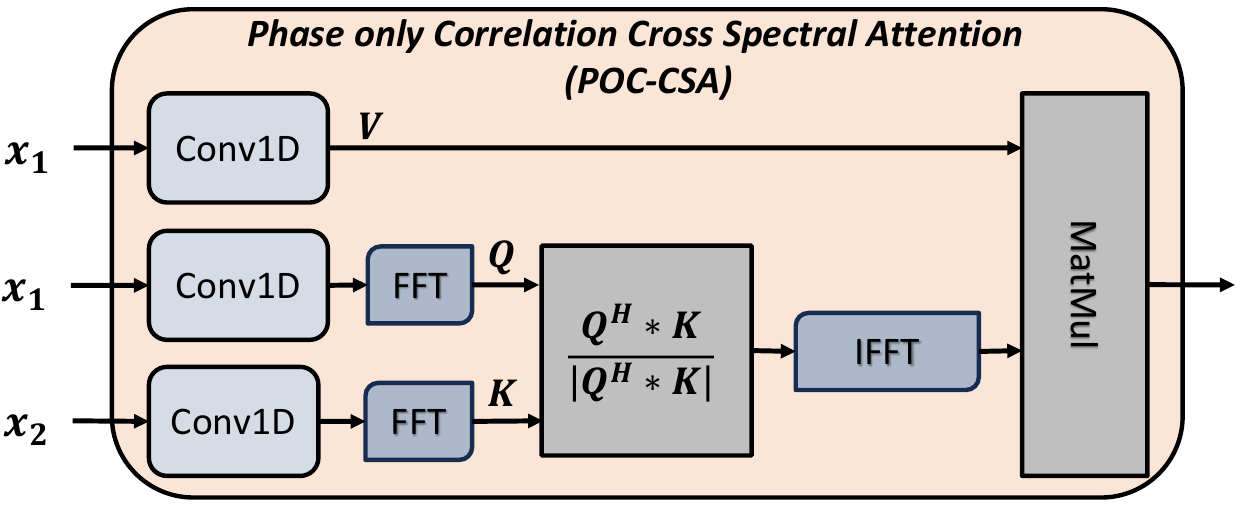}
\caption{Schematic diagram of POC-CA($x_1,\,x_2)$), where $x_1$ = channel input value and $x_2$ = cross-channel input; $Q^H$ represents conjugate transpose of $Q$.}
\label{fig:poc-csa}
\end{figure}

\subsubsection{Patch Sequencing} This step converts the vein pattern images to the sequences of token embeddings \cite{vit}. The patching step uses 2D convolution to convert the input image to a sequence of 2D patches of size  $(d_{p} \times d_{p})$. If the resulting number of patches is $N_{patches}$, one batch of input images is converted into $(b\times N_{e}\times d_{p} \times d_{p})$, where $b$ denotes the batch size and $N_e$ denotes the embedding size or number of patches. Now, each 2D patch in the flattened sequence is converted to 1D sequence of size ($d_{s} \times 1$), where $d_{s}$ = $d_{p} \times d_{p}$. The final flattened patches of shape $b\times N_{e}\times d_{s}$ are the input to each channel of the cross-spectral encoder.

\subsubsection{Cross-Attention Encoder} The encoder block consists of a dual-channel cross-spectral attention mechanism to accept flattened patches of the forehead and periocular channels independently and extract person-specific features. Each channel comprises a multi-head POC-CA, followed by a non-linear feed-forward block.

\subsubsection{Multi-Head POC-based cross-attention} The POC-based cross-attention block consists of multiple POC-CA heads in parallel. The basic principle behind POC-based cross-attention is to capture the phase correlation between the two biometric traits and multiply it by the corresponding channel value for a subject-specific transformation.  The schematic diagram of a POC-CA head is shown in Fig. \ref{fig:poc-csa}.  Each head takes three inputs: Query ($Q$), Key ($K$), and Value ($V$). Let $x_1$ and $x_2$ represent the inputs to channel-1 and channel-2, respectively. The multi-head POC-CA in the first channel computes cross-attention of $(x_1,\, x_2)$, while the multi-head POC-CA in the other channel computes cross-spectral attention of $(x_2,\, x_1)$.

For the computation of cross-attention of $(x_1,\, x_2)$, $V$ and $Q$ are calculated using $x_1$, while $K$ is calculated using $x_2$ as shown in equations (\ref{eq:v}), (\ref{eq:q}), and (\ref{eq:k}). 
\begin{eqnarray}
    V &=& f^{h}(x_1) \label{eq:v} \\
    Q &=& \mathscr{F}\left(f^{h}(x_1)\right)\label{eq:q}\\
    K &=& \mathscr{F}\left(f^{h}(x_2)\right)\label{eq:k}
\end{eqnarray}
where $f^{h}(.)$ represents 1D convolutional function with kernel size of $h$ and $\mathscr{F}(.)$ represents FFT. The convolutional function's kernel size ($h$) is chosen such that $h*$number of heads =  $N_{patch}$. The cross-attention score that represents phase-only information is obtained between $K$ and $Q$ as shown in Eq (\ref{eq:poc_cs}).
\begin{equation}\label{eq:poc_cs}
    \mathcal{M}_{p} = \frac{Q^H*K}{|Q^H*K|} 
\end{equation}
where, $Q^H$ denotes conjugate transpose of $Q$. The inverse fast Fourier transform of $\mathcal{M}_{p}$ represents the POC between $Q$ and $K$. The division by the magnitude term ensures that the term $\mathcal{M}_{p}$ represents phase only and rejects the effect of intensity. Therefore, the obtained POC represents the structural and geometrical map between the two inputs (the periocular and the forehead vein patterns), which carries unique information for the subject identification. Therefore, the real part of the POC obtained in Eq (\ref{eq:poc_cs}) is termed the attention score and is multiplied by $V$ to compute the final attention value ($\mathcal{M}$) as shown in Eq. (\ref{eq:attnOutput}).
\begin{equation}\label{eq:attnOutput}
    \mathcal{M}\; = \; V*SoftMax(\mathscr{R}(\mathscr{F}^{-1}(\mathcal{M}_{p})).
\end{equation}
where $\mathscr{R}(.)$ represents the real part of the complex output of the inverse transform $\mathscr{F}^{-1}(.)$. The size of the attention heads is selected such that the concatenated output of the multi-head block is the same as the flattened patch size ($b\times N_{e}\times d_{s}$). Therefore, no projection layer is required for combining the outputs of the multiple heads.  If  $\mathcal{M}_1$, $\mathcal{M}_2$, ... $\mathcal{M}_n$ are the outputs of $n$ heads, the final output of the multi-head POC-CA can be obtained by concatenating along the embedding axis using Eq. (\ref{eq:conc}):
\begin{equation}\label{eq:conc}
    \mathcal{M}_{o} = \text{Concat}(\mathcal{M}_1,\; \mathcal{M}_2,\; \cdots,\; \mathcal{M}_n)
\end{equation}

\subsubsection{Feed-Forward Layer} The output of the multi-head attention block is now added to the corresponding channel input (skip connection) and passed through layer normalization for training stabilization. Now, the normalized output is further transformed by a two-layer feed-forward neural network. The two layers of the MLP block sandwich a nonlinear activation layer that adds nonlinearity in the higher-dimensional space. Overall, this block transforms the attention features suitable for final classification.

\subsubsection{Classification Head} \label{sec:clshead} Finally, feature outputs from the dual-channel encoder are concatenated along the embedding (feature) dimension. The concatenated features are processed by a classifier head designed to transform the feature outputs to the final subject identification. The feature output of each encoder channel has a shape of $b\times d_h$. The concatenated features get a shape of $b\times 2d_h$. Therefore, the classifier head has the first linear layer with the number of nodes equal to $2d_h$, followed by a nonlinear transformation using a Parametric Rectified Linear Units (PReLU) activation layer, and the last softmax layer having nodes equal to the number of classes or subjects in the dataset.

\section{Experimental Results and Discussion} \label{sec:experi}
This section presents the evaluation of the proposed POC-ViT framework using an extended version of the Forehead Vein Pattern and Periocular Biometric Pattern (FSVP-PBP) database. The database consists of 25 frames per subject, captured for 350 individuals (aged 21–42) using an NIR camera in a controlled environment (illumination between 20–40 Lux). Subjects maintained a 4–6 cm distance from the camera, and facial surfaces were cleaned prior to data collection to ensure image clarity.

To evaluate the robustness and performance of the proposed approach, the dataset of 8750 images was split into 80\% for training and 20\% for testing using random sampling. Two cases were considered for training and validation:

\begin{itemize}
    \item \textbf{Case (i): Without Preprocessing} --- The raw NIR images were directly fed into the POC-ViT network. While NIR imaging often introduces luminance-related noise, the POC-based cross-spectral attention in POC-ViT effectively extracts robust phase information, mitigating luminance-induced degradation.

    \item \textbf{Case (ii): With Adaptive Tann-Triggs Preprocessing} --- Here, the images were preprocessed using the adaptive Tann-Triggs algorithm, which dynamically adjusts preprocessing parameters based on image statistics to enhance the visibility of vein patterns.
\end{itemize}

To quantify the performance of the proposed framework, standard evaluation metrics were adopted, including Classification Accuracy (CA) \cite{bhattacharya2019multi}, Equal Error Rate (EER) \cite{phillips2000feret}, True Acceptance Rate (TAR) at FAR = 0.1\% \cite{phillips2000feret}, and training dynamics visualized through epoch-wise accuracy trends. These metrics were consistently applied across both preprocessing scenarios. The resulting performance comparisons, EER plots, and training/validation trends collectively demonstrate that the POC-ViT framework achieves high classification accuracy, low error rates, and robust generalization capability, with marginal improvements observed when using the adaptive Tann-Triggs preprocessing algorithm.


\subsection{Implementation Details}  
The proposed POC-ViT, as shown in Fig. \ref{fig:proposed} is implemented using the PyTorch library of Python on the Linux platform. The model's hyper-parameters were selected as follows: number of CS-Encoder blocks = $6$, number of heads in multi-head attention blocks = $8$, input image size = $256$, patch size = $16\times16$, the embedding dimension = $512$, sequence size = $16\times 16 = 256$, and number of hidden sizes for the non-linear feed-forward block = $2*$embedding dimension. Therefore, the output of the last encoder at each channel has a shape of $b\times 256\times 512$, where $b$ represents the batch size. The final output features of each channel are obtained by taking the mean along the sequence dimension ($256$), which ultimately converts to the shape of $b\times 512$. Both feature outputs are concatenated along the feature dimension to obtain a feature size of $b \times 1024$. 

The concatenated output of the encoder block is passed through a custom classifier, designed to accept input features of size $1024$ with one intermediate layer to transform the feature size successively from $1024\; \longrightarrow\; 512\; \longrightarrow\; 350$, where $350$ is the number of classes in the dataset.  The intermediate layer having $512$ hidden nodes is activated using a learnable parametric activation function, called PReLu, described in Section \ref{sec:clshead}. 

\begin{figure}[!ht]
\centering
\includegraphics[width=\columnwidth]{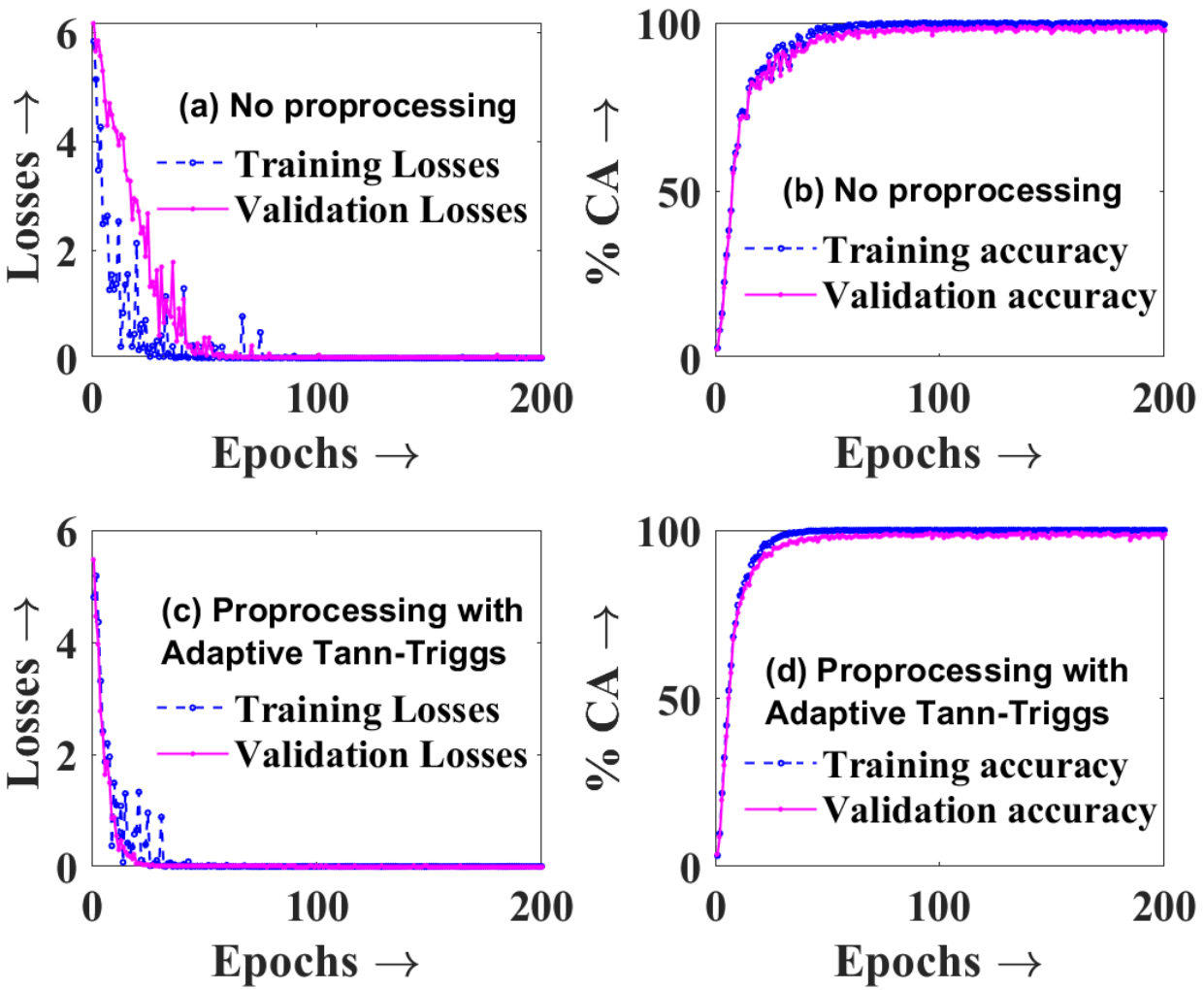}
\caption{Training and validation statistical curve for the proposed POC-ViT: (a) training and validation losses }
\label{fig:tr_log}
\end{figure}

The weights of the model are initialized using Xavier initialization. The model is trained using the training dataset under the two cases as described above for $200$ epochs on the \textit{NVIDIA Tesla V100} GPU at SMSS Lab, Department of Mechanical Engineering, IIT Kanpur. The training statistics curve in terms of training accuracy and validation accuracy is shown in Fig. \ref{fig:tr_log}

\begin{figure}[!ht]
\centering
\includegraphics[width=\columnwidth]{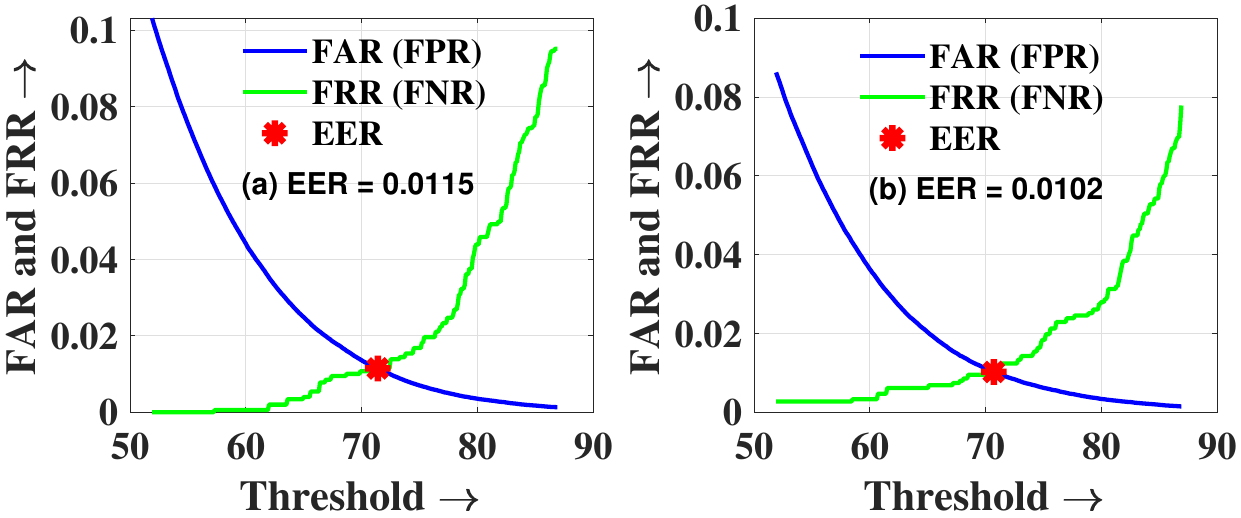}
\caption{Computation EER using FAR curve and FRR curve for POC-ViT on the validation data with: (a) no preprocessing and (b) preprocessing by adaptive Tann-Triggs algorithm.}
\label{fig:far_frr_eer}
\end{figure}


\subsection{Comparison with State-of-the-art Methods} 
For further validation of the proposed framework for biometric authentication, this section presents a comparative analysis of the performance of various algorithms on the aforementioned extended FSVP dataset. The study encompassed a range of recent methods for comparison: VGG-16 \cite{vggnet}, ResNet-50 \cite{resnet50}, GoogleNet \cite{googlenet}, PV-CNN \cite{das2018convolutional}, 2D FV-ViT \cite{FV-ViT},  FV-LT \cite{FV-LT}, VPCFormer \cite{VPCFormer}, and ABLSTM-TSVP \cite{blstm}. All these algorithms accept single-channel input only, but the FSVP dataset contains forehead vein patterns and particular images captured by the NIR camera. Therefore, images of the two biometric traits (forehead and periocular) were concatenated along the suitable dimension to make single-channel input for each subject from the train and the test samples. Each model was trained for 100 epochs and evaluated on the test dataset. The performances in terms of CA, EER, and TAR have been presented in Tables \ref{tab:res_noPre} and \ref{tab:res_att} for the cases of no preprocessing and preprocessing with the adaptive Tann-Triggs algorithm. It can be observed that the proposed framework demonstrated the highest performance with near-perfect classification, evident from the highest values of CA's and TAR's, and lowest values of EER's in both cases. The comparison of the results by the proposed framework under the two cases in Tables \ref{tab:res_noPre} and \ref{tab:res_att} reveals that the POC-based cross-spectral attention mechanism is sufficiently robust to capture the structural pattern from the NIR camera-based raw images. However, the application of the adaptive Tann-Triggs algorithm for preprocessing the vein patterns marginally enhances the performance of the proposed POC-ViT framework. Very low EER value and high TAR signify that the proposed model offers a highly reliable security system with a very low failure rate and high acceptance of legitimate users only.


\begin{table}[!ht]
  \centering
  \caption{Performance comparisons for the dataset with no preprocessing}
  \resizebox{0.9\columnwidth}{!} {
{
\begin{tabular}{|l|c|c|c|}
\hline
\textbf{Method} & \textbf{Accuracy} & \textbf{EER} & \textbf{\makecell{TAR\\@FAR=0.1\%}} \\
\hline
{VGG16} \cite{vggnet} & 91.6	& 2.512	& 58.40\\
\hline
{ResNet50} \cite{resnet50} & 92.7  & 1.806 & 82.60 \\
\hline 
{GoogleNet} \cite{googlenet} & 93.5  & 0.986 & 78.80 \\
\hline 
{PV-CNN} \cite{das2018convolutional} & 88.8  & 2.370 & 62.05 \\
\hline 
{2D-FV-ViT} \cite{FV-ViT} & 94.6  & 1.875 & 87.60 \\
\hline 
{FV-LT} \cite{FV-LT} & 95.0  & 1.056 & 90.60 \\
\hline 
{VPCFormer} \cite{VPCFormer} & 96.8  & 0.984 & 95.50 \\
\hline 
{ABLSTM-TSVT} \cite{blstm} & 96.4  & 0.950 & 97.50 \\
\hline 
\textbf{POC-ViT (Proposed)} & \textbf{98.3}  & \textbf{0.012} & \textbf{97.80} \\
\hline
\end{tabular}}}
\label{tab:res_noPre}%
\end{table}%

\begin{table}[!ht]
  \centering
  \caption{Performance comparisons for the dataset with preprocessing using adaptive Tann-Triggs}
  \resizebox{0.9\columnwidth}{!} {
{
    \begin{tabular}{|l|c|c|c|}
    \hline
    \textbf{Method} & \textbf{Accuracy} & \textbf{EER} & \textbf{\makecell{TAR\\@FAR=0.1\%}} \\
    \hline
    \textbf{VGG16} \cite{vgg16} & 94.3  & 2.512 & 58.40 \\
    \hline
    \textbf{ResNet50} \cite{resnet50} & 94.5  & 1.806 & 82.60 \\
    \hline
    \textbf{GoogleNet} \cite{googlenet} & 95.5  & 0.986 & 78.80 \\
    \hline
    \textbf{PV-CNN} \cite{das2018convolutional} & 90.3  & 2.370  & 62.05 \\
    \hline
    \textbf{2D-FV-ViT} \cite{FV-ViT} & 95.34 & 1.875 & 87.60 \\
    \hline
    \textbf{FV-LT} \cite{FV-LT} & 97.5  & 1.056 & 90.60 \\
    \hline
    \textbf{VPCFormer}\cite{VPCFormer} & 98.0    & 0.984 & 95.50 \\
    \hline
    \textbf{ABLSTM-TSVT} \cite{blstm} & 97.5  & 0.950  & 97.50 \\
    \hline
    \textbf{POC-ViT (Proposed)} & \textbf{98.8}  & \textbf{0.010} & \textbf{98.50} \\
    \hline
    \end{tabular}}}
  \label{tab:res_att}%
\end{table}%

\subsection{Comparison with State-of-the-art Methods with other Biometric modalities}
\begin{table*}[!ht]
\centering
\caption{Comarisons with State-of-the-art methods on various modalities of biometric systems.}
\label{systemcomparison1}
\resizebox{0.9\textwidth}{!}
{
\begin{tabular}{|c|c|c|c|c|c|c|}
\hline
\textbf{Reference} & \textbf{Database} & \textbf{\#Samples} &  \textbf{\#Subjects} & \textbf{Method} & \textbf{Accuracy} & \textbf{EER} \\ \hline

\cite{asaari2014fusion} & FVP-USM & 492 & 123 & BLPOC, WCCD & - & 2.34 \\ \hline

\cite{kang2014contact} & CASIA-MPID & 1260 & 105  & RootSIFT + LBP  & - & 0.996 \\ \hline

\cite{oh2014combining} & SRP+PBP & 1877 & - & MLBP & - & 3.26 \\ \hline

\cite{santos2015fusing} & CSIP & 3732 & 92 & \begin{tabular}[c]{@{}c@{}}LBP, HOG, ULBP, \\ GIST, SIFT, Iriscode\end{tabular} & 93.50 & 0.145 \\ \hline

\cite{lin2015bimodal} & PPP+DHVP & 3000 & 100 & DWT+IDWT+SVM & 98.80 & 1.38 \\ \hline

\multirow{2}{*}{\cite{qin2017deep}} & FVP-USM & 492 & 123 & \multirow{2}{*}{\begin{tabular}[c]{@{}c@{}}DCNN \end{tabular}} & 69.33 & 0.80 \\ 
\cline{2-4} \cline{6-7} 
 & FVP-HKPU & 302 & 156 &  & 84.59 & 2.33 \\ \hline

\cite{wang2017quality} & DHVP & 500 & - & \makecell{Discriminative LBP\\with Chi-Square} & 95.59 & 0.08 \\ \hline

\cite{ma2017palm} & \makecell{CASIA-MS \\PalmprintV1} & 7200 & 100 & \makecell{Adaptive Gabor Filter\\with Hamming Filter} & - & 0.12 \\ \hline

\cite{ahmad2019lightweight} & \makecell{PolyU Multispectral\\Palmprint database} & 6000 & 250 & \makecell{Wave Atom transform\\ with Hamming Filter} & 95.26 & 1.98 \\ \hline

\cite{daas2020multimodal} & SUHM-FV + HKPolyU FKP & 3816 & 106 & DCNN & 98.31 & 0.29 \\ \hline

\multirow{2}{*}{\cite{noh2020finger}} & HKPolyU-FI & 1872 & 156 & \multirow{2}{*}{DenseNet-161} & - & 0.05 \\ \cline{2-4} \cline{6-7}

 & SUHM-FV & 3816 & 106 &  & - & 1.65 \\ \hline

\cite{li2021local} & FP+FV+FKP  & 17500 & 500 & LC-CNN + SVM  & 99.90 & 1.21 \\ \hline

\cite{bhattacharya2022portable} & FSVP+PBP & 1600 & 200 & DCNN & 98.60 & 0.08 \\ \hline

\cite{huang2023multimodal} & FP+FV & 1800 & 256 & FAB-Net & 98.84 & 0.17 \\ \hline

\textbf{Proposed} & \textbf{FSVP+PBP} & \textbf{8750} & \textbf{350} & \textbf{POC-ViT} & \textbf{98.8} & \textbf{0.010} \\ \hline
\end{tabular}
}
\end{table*}
The proposed POC-ViT framework for biometric authentication, trained on the extended database of the FSVP-PBP dataset having 350 subjects, demonstrates superior performance in comparison to state-of-the-art (SOA) biometric systems trained on the same dataset. The POC-ViT model, along with the adaptive Tan-Triggs preprocessing, achieves a classification accuracy of $98.8\%$ and a very low EER of $0.010$, significantly outperforming other methods. This section provides a comparative analysis of other modality of biometric systems: finger vein (FV), dorsal hand vein patterns (DHVP), and dynamic palm vein matching (DPVM), iris and periocular patterns (IP + PBP), sclera and periocular patterns (SRP + PBP), and palm print with dorsal hand vein patterns (PPP + DHVP) as summarised in Table \ref{systemcomparison1}. For example, the DPVM method on the PolyU database achieves a classification accuracy of $95.26\%$ with an EER of $1.98$, while another DPVM method on the CASIA database achieves an EER of $0.12$. Furthermore, multimodal systems combining various patterns, such as IP + PBP, SRP + PBP, and PPP + DHVP, also fall short in comparison. The PPP + DHVP system achieves a classification accuracy of $98.80\%$ with an EER of $1.38$. These results underscore the effectiveness of the proposed POC-ViT framework in achieving unprecedented accuracy and reliability for biometric authentication, making it a promising solution for secure and non-intrusive identification technologies.

\subsection{Discussion on Deployability}
\label{DiscussionOnDeployment} 
The computational complexity of the proposed framework is compared with the state-of-the-art methods in terms of the floating point operations (FLOPs) and the number of trainable parameters in Table \ref{tab:comp-cost}. The value of FLOPs for each model is calculated based on compute operations required for the forward pass using one batch of the dataset with the image size of $3\times 256\times256$. Since the proposed framework is a dual architecture to handle the dual biometric traits simultaneously, the two forehead and the periocular portions are concatenated before feeding to the state-of-the-art models for fair comparison. It can be observed that the value of the FLOPs for the proposed model is the lowest, even though the number of trainable parameters for the GoogleNet, VPCFormer, and ResNet50 are lower compared to our model. The FLOPs for these models become higher because of the concatenated input of the dual biometric traits. Furthermore, the number of CS-Encoder layers in the proposed framework is $6$, which is half of the $12$ layers in the standard ViT model \cite{FV-ViT}. Also, the size of the embedding dimension is 512, instead of 768 in the standard ViT model \cite{FV-ViT}. Therefore, the proposed framework is a lightweight model with the best performance and robustness against illumination variation. The lowest EER and highest TAR justify the best reliability of the authentication by the proposed framework. Therefore, the proposed framework can be a new benchmark for the real-time deployment of touchless biometric authentication.

\begin{table}[!ht]
\centering
\caption{Comparing the FLOPs and the trainable parameters.}
\label{tab:comp-cost}
\resizebox{0.8\columnwidth}{!}
{
\begin{tabular}{|l|c|c|}
\hline
\textbf{Method} & \makecell{\textbf{FLOPs}\\\textbf{(Billion)}} & \makecell{\textbf{Parameter}\\\textbf{(Million)}} \\ \hline
{VGG16} \cite{vgg16} & 643.36 & 135.69 \\
\hline
{ResNet50} \cite{resnet50} & 173.16 & 24.22 \\
\hline
{GoogleNet} \cite{googlenet} & 63.26 & 5.95 \\
\hline
{PV-CNN} \cite{das2018convolutional} & 3442.34 & 59.35 \\
\hline
{2D-FV-ViT} \cite{FV-ViT} & 111.07 & 86.36 \\
\hline
{FV-LT} \cite{FV-LT} & 42.53 & 31.60 \\
\hline
{VPCFormer} \cite{VPCFormer} & 26.92 & 21.72 \\
\hline
{ABLSTM-TSVT} \cite{blstm} & 77.91 & 100.09 \\
\hline
\textbf{POC-ViT (Proposed)} & \textbf{13.32} & \textbf{26.46} \\ \hline
\end{tabular}
}
\end{table}

\section{Conclusions}
\label{sec:conclusion}  
In this paper, we introduced a novel POC-ViT framework for biometric authentication that utilizes the concept of POC to compute the attention score in a multihead cross-attention mechanism. The proposed Phase-Only Correlation-based Cross-Attention (POC-CA) mechanism effectively captures phase-based structural features from the two biometric traits, ensuring robustness against variations in illumination and magnitude. A key innovation of the proposed approach lies in the ability of the POC-CA mechanism to extract essential biometric features using the forehead and the periocular images for accurate and precise subject identification. Further, the proposed adaptive Tann-Triggs algorithm for preprocessing the NIR images enhances the reliability of the POC-ViT framework for biometric authentication.  The evaluation of the proposed framework on the extended FSVP-PBP database having 350 subjects demonstrates its superiority over state-of-the-art algorithms. The remarkable performance of the proposed framework underscores its potential as a robust and reliable solution for non-intrusive and touchless identity verification. Our findings not only contribute to the advancement of biometric authentication technologies but also open avenues for further research in exploring the capabilities of cross-spectral vision transformers in other domains of pattern recognition and image analysis.

\ifCLASSOPTIONcaptionsoff
  \newpage
\fi

\section*{Acknowledgements} 
The Authors are sincerely thankful to volunteers from the Hijli Co-operative Society, Prembazar, Kharagpur and the students of Indian Institute of Technology, Kharagpur, W.B. India for their participation and consent to capture NIR images of the forehead and periocular portions. We are committed to upholding the highest ethical standards, ensuring all data are maintained anonymously.

\section*{Author Credit Declaration}
 \textbf{Arun K. Sharma:} Conceived and formulated the idea of POC-based cross-spectral attention mechanism and the cross-spectral ViT for biometric authentication. Code Implementation, dataset preprocessing, training, and validation of the POC-ViT model. Wrote and critically examined the manuscript.  \textbf{Shubhobrata Bhattachary:} Dataset collection and curation.  Wrote and critically examined the manuscript. \textbf{Motahar Reza:} Partially involved in the mathematical formulation of the POC-ViT model. Wrote and critically examined the manuscript. \textbf{Bishakh Bhattacharya:} Technical Supervision. Wrote and critically reviewed the manuscript. Funding for research work.


\bibliographystyle{IEEEtran.bst}
\bibliography{Reference.bib}

\begin{IEEEbiography}[{\includegraphics[width=1in,height=1.5in,clip,keepaspectratio]{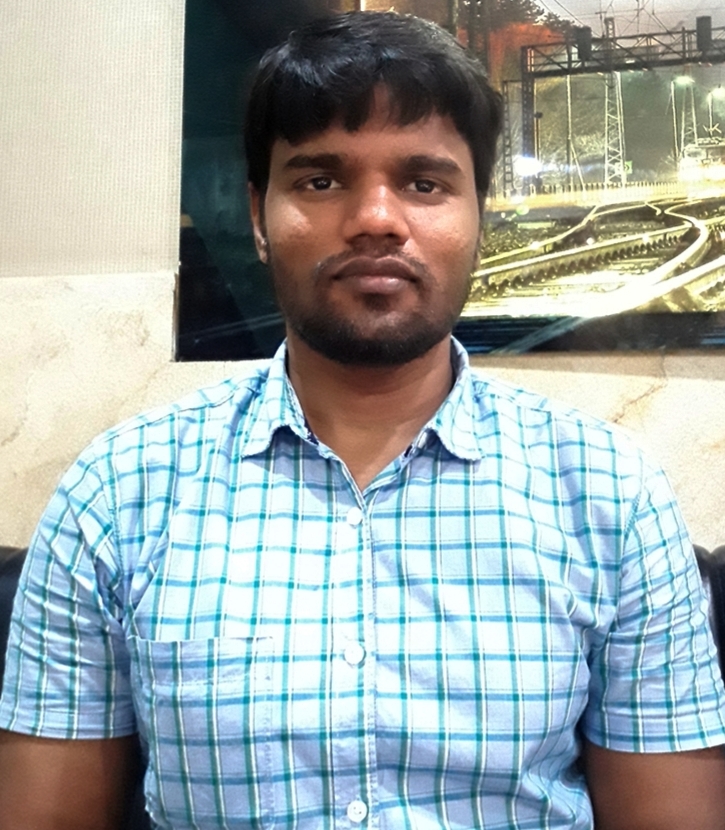}}]{Arun K. Sharma} received his B.Tech. in Electrical Engineering from BIT Sindri Dhanbad and M.Tech. in Instrumentation from Indian Institute of Technology, Kharagpur. He earned his PhD in Electrical Engineering from Indian Institute of Technology, Kanpur. He is currently working as a post-doctoral researcher at SMSS Lab, IIT Kanpur.  His research interests are signal processing, deep learning for pattern recognition and analysis, and AI in advanced engineering applications. He has been a reviewer for several reputed journals, such as the IEEE Transactions on Aerospace and Electronic Systems (TAES), the IEEE Transactions on Fuzzy Systems, and the IEEE Computational Intelligence Magazine.
\end{IEEEbiography}
\begin{IEEEbiography}[{\includegraphics[width=1in,height=1.25in,clip,keepaspectratio]{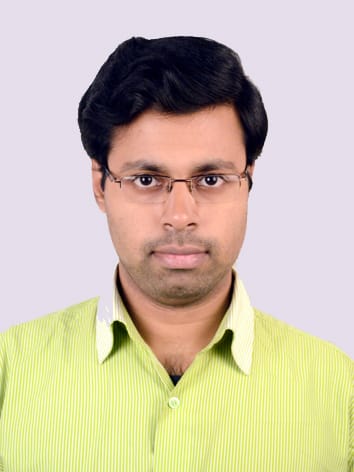}}]{Shubhobrata Bhattacharya} received his B.Tech. degree in Electrical Engineering from Kolaghat Engineering College. He obtained his M.Tech. and Ph.D. degrees from the Indian Institute of Technology Kharagpur. His Ph.D. research was focused on Face Recognition and Biometric Applications, contributing to advancements in robust identification systems. His present research interests include computer vision, machine learning, linear algebra, signal processing, and image processing, with a special emphasis on deep learning-based quality assessment and recognition frameworks. He actively serves as a reviewer for reputed journals such as Expert Systems, Pattern Recognition, Pattern Recognition Letters, IEEE Transactions on Instrumentation and Measurement, and IEEE Transactions on Circuits and Systems for Video Technology.
\end{IEEEbiography}
\begin{IEEEbiography}
[{\includegraphics[width=1in,height=1.25in,clip,keepaspectratio]{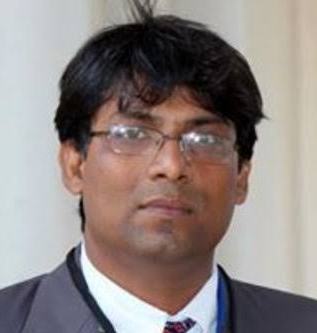}}] {Motahar Reza} received his B.Sc. in Mathematics (Honors) from Calcutta University, his M.Sc. in Applied Mathematics from Jadavpur University, Kolkata, and his M.Tech in Data Science and Engineering from BITS Pilani. He earned his Ph.D. from the Indian Institute of Technology, Kharagpur. He was awarded the BOYSCAST Fellowship by the DST, Government of India, for research at LSTM, Friedrich-Alexander-Universität Erlangen-Nürnberg, Germany. In 2008, he received the Orissa Young Scientist Award. He has secured numerous research and other grants from SERB, DST, AICTE, DRDO, CSIR, and TEQIP-II BPUT under various schemes. His current research interests include Machine Learning, Data Science, Parallel Algorithms, and Computational Science. He is currently an Associate Professor at GITAM University, Hyderabad, India.
\end{IEEEbiography}

\begin{IEEEbiography}
[{\includegraphics[width=1in,height=1.25in,clip,keepaspectratio]{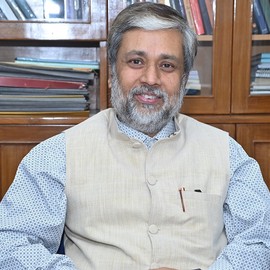}}] {Bishakh Bhattacharya} is currently Professor, HAG, in the department of Mechanical Engineering at IIT Kanpur. Prof. Bhattacharya is also working as the Prof. in Charge for the development of Indian Institute of Skills and the CDSTAR – an ambitious endeavor of the Govt. of India to develop new generation advanced skilling centre.  Prof. Bhattacharya hold formerly the HAL Chair and  Dr. and Mrs. G. D. Mehta Chair in the Institute. His current research interest is Physically Intelligent Robots, Child-Robot Interaction, Vibration Control by Active and Passive Smart materials, Active Shape Control of Flexible Systems, Structural Health Monitoring and Intelligent System Design.
He has been involved in modeling and development of hybrid composite laminate activated by smart materials like Terfenol-D alloy and PZT and found its application in controlling vibration of flexible rotating links like helicopter rotor. 
Prof. Bhattacharya lead the development of may national projects including the development of Pipe Health Monitoring Robot for GAIL, Substation Inspection Robot for Powergrid, Compressed Air-based Cargo Hyperloop for CMPDI, a coordinated UGV and UAV for Powerline Monitoring by CPRI, Semi-autonomous Stubble Harvesters and Robots for Child-Robot Interaction. He has more than 30 patents and published 120 international journal papers. He is the recipient of GCRF visiting fellow at Swansea University, UKIERI Visiting fellow at the University of Sheffield, DST-JSPS visiting fellow at the Kyushu Institute of Technology, SAKURA fellow at the Yokohama National University, JAPAN.
\end{IEEEbiography}

\end{document}